\documentclass[10pt,twocolumn,letterpaper]{article}

\usepackage{iccv}
\usepackage{times}
\usepackage{epsfig}
\usepackage{graphicx}
\usepackage{algorithm}
\usepackage{algorithmic}
\usepackage{bbm}
\usepackage{mathrsfs}
\usepackage{multirow}
\usepackage{cite}
\usepackage{amsmath,amssymb,amsfonts}
\usepackage{textcomp}
\usepackage{listings}
\usepackage{caption}
\usepackage{subcaption}

\usepackage[pagebackref=true,breaklinks=true,letterpaper=true,colorlinks,bookmarks=false]{hyperref}

\iccvfinalcopy 

\def\iccvPaperID{***} 

\begin{document}

\title{SPGen: Stochastic scanpath generation for paintings using unsupervised domain adaptation}

\author{Mohamed Amine KERKOURI{$^{1}$},~ marouane Tliba{$^{2}$},~ Aladine Chetouani{$^{2}$},~ Alessandro bruno{$^{3}$}\\\vspace{-8pt}{\small~}\\
{$^{1}$}F-Initiatives;~~~ {$^{2}$}Université Sorbonne Paris Nord; ~~~ {$^{2}$}IULM University;\\
}
\maketitle

\begin{abstract}
Understanding \textit{human visual attention} is key to preserving \textit{cultural heritage}. We introduce SPGen, a novel \textit{deep learning} model to predict \textit{scanpaths}: the sequence of eye movements—when viewers observe paintings.

Our architecture uses a \textit{Fully Convolutional Neural Network (FCNN)} with differentiable fixation selection and learnable Gaussian priors to simulate natural viewing biases. To address the domain gap between photographs and artworks, we employ \textit{unsupervised domain adaptation} via a gradient reversal layer, allowing the model to transfer knowledge from natural scenes to paintings. Furthermore, a random noise sampler models the inherent stochasticity of \textit{eye-tracking} data.

Extensive testing shows SPGen outperforms existing methods, offering a powerful tool to analyze \textit{gaze behavior} and advance the preservation and appreciation of artistic treasures.
\end{abstract}

\section{Introduction}
\label{sec:intro}

Cultural heritage encompasses the habits, traditions, artifacts, and artistic expressions that a community creates and passes down through generations \cite{HARTMANN2020369}. This collective heritage forms an essential part of the community's identity, representing a treasure trove of their shared values and experiences that must be safeguarded for posterity.

One significant aspect of this cultural heritage is paintings, which hold a prominent place in the artistic legacy of humanity. A painting is a form of visual art created by applying paint to a surface, and it showcases the artist's skill and creativity in conveying ideas and emotions through images \footnote{\url{https://dictionary.cambridge.org/dictionary/english/painting}}. Paintings serve as visual narratives, offering glimpses into historical events, societal norms, and the artistic inclinations of the time and the artists themselves.

Paintings encapsulate the aesthetics of a particular era and the cultural, political, and philosophical underpinnings that shaped the artists' perspectives. By cherishing and protecting these artistic treasures, we can better grasp the nuances of our past, fostering a deeper appreciation for the diversity and richness of human creativity across time and space.

Preserving and comprehending such artistic heritage is paramount, as it contributes significantly to our understanding of human history and the evolution of societies.

The understanding and comprehension of this form of media are intricately connected to the cognitive and perceptual mechanisms of the human visual system (HVS) \cite{von1925helmholtz} \cite{CVD_ETRA}. Human eyes possess remarkable capabilities, with a resolution of over 500 megapixels (MP), allowing them to capture vast visual information. Remarkably, the human visual system exhibits an extraordinary capacity, whereby the human eyes are capable of receiving an estimated influx of approximately 10 Gb/s of visual information. \cite{brain}.


Given the magnitude of data requiring processing and the limited disponibility of cognitive resources, the human visual system (HVS) necessitates an efficient mechanism for filtering the most relevant information by focusing on the important spacial regions within a visual scene, while concurrently disregarding the less significant areas from a perceptual standpoint. This mechanism is called visual attention.

In greater detail, the process of eye movements plays a crucial role in filtering non-relevant information by selectively attending to specific regions of the visual field while disregarding others. This dynamic mechanism effectively unburdens the cognitive load on the human visual system and enhances the overall efficiency of the visual perception system.

Findings revealed that during the early stages of image observation, low-level visual features, such as color, edges, texture, intensity, and contrast, are the primary factors that attract eye movements \cite{feature_integration}. These fundamental visual characteristics draw immediate attention and guide the initial exploration of the visual scene.

In contrast, during a later stage of the visual experience, semantic visual features become prominent in directing eye movements. These higher-level features encompass meaningful elements, including faces, text, humans, and objects, which hold considerable significance in visual perception and cognition. As the observer becomes more engaged with the image and comprehends its content, attention shifts towards these semantically meaningful aspects.

Thus visual attention can be classified based on the neural processing pathways it employs \cite{attetnion_types}. "Bottom-up Attention" is primarily driven by external stimuli features, such as the low-level visual features mentioned earlier. In this type of attention, the intrinsic properties and characteristics of the visual stimuli dominate the mechanism, aligning with the theory that attention is a pre-processing operation. Bottom-up Attention operates unconsciously, rapidly, and instinctively, directing our focus to salient and noteworthy elements in the visual environment.

On the other hand, "Top-down Attention" is influenced by higher-level cognitive processes originating from the prefrontal cortex and long-term memory. This type of attention is guided by internal factors, such as prior knowledge, expectations, and goals, rather than solely relying on the properties of the stimuli. Top-down Attention corresponds to the theory that attention is a post-processing operation, as it involves deliberate and intentional allocation of focus. This mechanism is characterized by its deliberate, slow, and premeditated nature, allowing us to selectively attend to specific aspects of the visual scene based on our objectives and cognitive strategies.

The aforementioned mechanism of attention can be further categorized into two types: 'covert' and 'overt' attention \cite{ZHANG2012183} \cite{Kulke_Atkinson_Braddick_covert_overt}. Covert attention refers to the allocation of attention without any accompanying eye movements. In this form of attention, the brain optimizes its processing activity in the visual cortex to prioritize specific regions of the visual field. It allows the observer to focus on relevant information without the need to physically shift their gaze.

In contrast, overt attention is closely linked to eye movements. It involves directing the fovea, the region of the retina with the highest visual acuity, towards regions of interest in the visual field. Overt attention, thus, aligns with the physical act of moving one's eyes to bring relevant areas into sharp focus. By doing so, the observer can extract more detailed information from the selected regions and further process the visual input with increased precision.

The distinction between covert and overt attention underscores the dynamic and adaptable nature of the human visual system. Covert attention allows for rapid and flexible shifting of focus within the visual scene without requiring explicit eye movements, while overt attention leverages eye movements to strategically examine specific areas in greater detail. These complementary attentional mechanisms enable humans to explore and comprehend complex visual environments efficiently, providing a crucial foundation for various cognitive processes.

Research studies in the field of visual attention predominantly emphasize the examination of measurable overt attention. To study overt attention, researchers commonly utilize eye-tracking technology, which captures and records the eye movements of viewers while they explore a particular visual stimulus. The data obtained from eye-trackers enables researchers to construct the gaze trajectory of individuals as they interact with the visual content.

The gaze trajectory comprises two primary components: fixation points and saccades. Fixation points represent moments when the eyes remain relatively still and focus on specific locations within the visual scene. During fixations, visual information is acquired and processed. On the other hand, saccades are rapid eye movements that occur between fixation points \cite{10.1145/1743666.1743717}. During saccades, no visual perception occurs as the eyes swiftly move from one location to another. This gaze trajectory is called scanpath and represents the main subjective measure of overt visual attention. 

Fixations made by multiple observers when presented with a specific stimulus, such as an image or a video sequence, are commonly depicted using fixation point maps. These maps offer a visual representation of the discretized distribution of gaze points across the stimulus. 

Smoothing the fixation point maps using a Gaussian kernel that corresponds to $1^\circ$ of visual angle is a common technique to achieve a smoother representation of eye movements \cite{KochUllman}. This smoothing technique yields a spatial distribution of eye movements and helps to reduce noise, and enhances the visualization of the overall pattern of attention across the stimulus.

A valuable outcome of smoothing the fixation point maps is the creation of saliency maps. These saliency maps provide a probability heat map wherein each pixel is assigned a normalized probability value indicating the likelihood of attracting the viewer's attention. Areas with higher probabilities on the saliency map are considered perceptually more relevant. Therefore, they are more likely to capture the observer's focus. 

Visual attention has become an important component that is used in  multiple fields ( i.e. technological, medical, educational, ... ) and several topics such as image quality assessment \cite{QAChetouaniICIP2018,Chetouani20EUSIPCO,Ilyass19NCAA,PR20Ilyass,SPIC20Chetouani}, image and video compression \cite{saliencyComp1}, image captioning and description \cite{saliencyCapt1}, image search and retrieval \cite{retrivalSal}, image enhancement for people with CVD (Colour Vision Deficiency) \cite{bruno2019image} \cite{CVD_ETRA}, saliency led painting restoration \cite{painting_restor} and so forth \cite{info19Hamidi} \cite{scanpath_MAE}.

Thus modeling the behavior of this mechanism holds accurately utmost significance and is a primary focus of interest \cite{satsal}, \cite{Tliba_2022_CVPR}.

\begin{figure*}
    \centering
    \includegraphics[width= 0.9\linewidth]{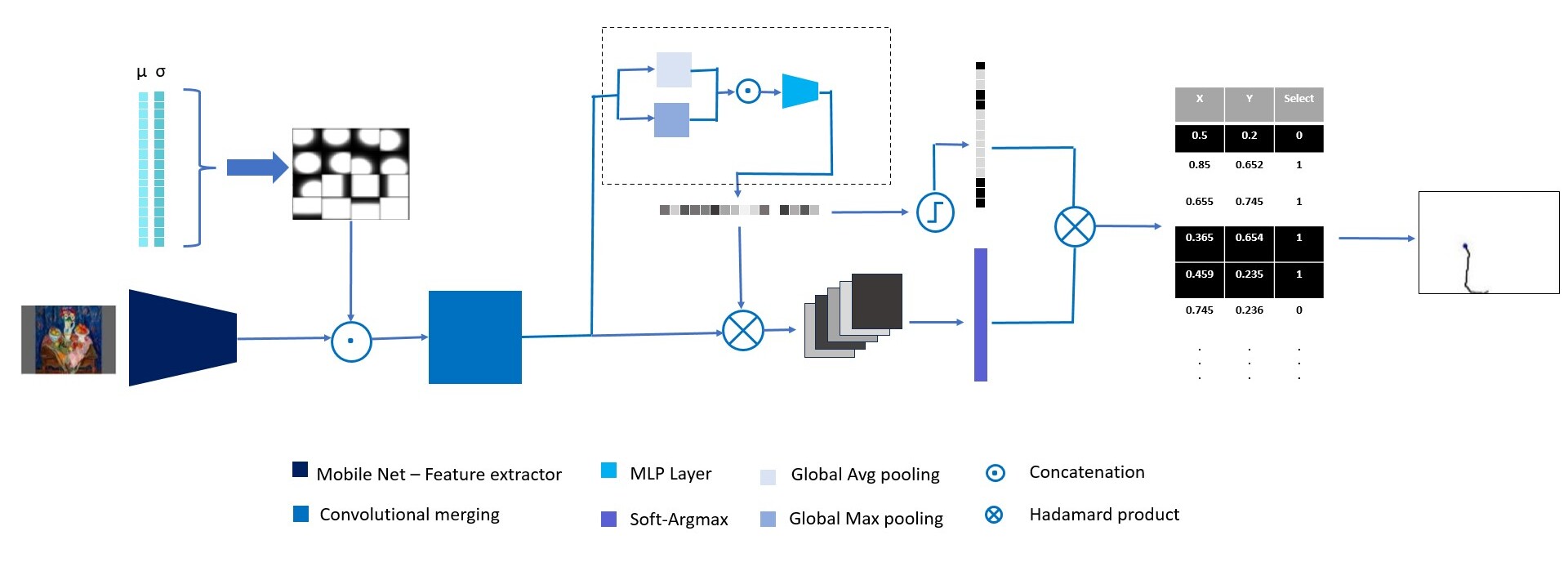}
    \caption{\label{figure:arch}General architecture of the proposed method.}
\end{figure*}

In this research paper, we introduce a novel and efficient approach for vector-to-sequence modeling, specifically focusing on converting images to scanpaths. Our method leverages fully Convolutional Neural Networks to achieve this task.

One of the challenges we address in our work is the presence of various biases in images originating from different domains, such as natural scenes, art, and synthetic data. These domain-specific biases, combined with subjective biases related to visual perception tasks, create noisy distributions that can hinder models from accurately capturing features specific to the images and relevant to the task.

We explore two types of domain adaptation to tackle this issue.
The first technique deals with formulating downstream task priors biases, while the second technique centers around generalizing feature extraction, also known as domain representation alignment \cite{alignment}. These techniques are applied across diverse data distribution domains to improve the prediction accuracy on  painting datasets.

Neural networks exhibit a deterministic behavior; which makes incorporating noise in them a challenge.
Incorporating noise in neural networks is a critical strategy due to its deterministic nature. 
Neural networks typically produce fixed outputs for a given input, which can limit the diversity in generated results, especially when dealing with stochastic and subjective phenomena like gaze scanpaths. Thus considering the stochastic and subjective nature of gaze scanpaths, we 
incorporate a random noise sampler that operates within the latent space of the representations.
This helps effectively manage the model's variability, which is controlled by a temperature parameter. The latter regulates the intensity of the noise by capturing the inherent uncertainty and subjectivity in human gaze behavior.

The main contributions of this work are summarized as follows: 
\begin{itemize}
  \item Propose an efficient deep learning model for scanpath prediction.
  \item Incorporate a selective module that enables the network to generate variable-length scanpaths.
  \item  Introduce a stochastic mechanism into the model to generate several scanpaths for the same stimuli.
  \item  Incorporate a task 
  domain-specific module that accounts for prior knowledge.  
  \item Enhance our model's generalization to painting images using unsupervised adversarial domain adaptation. This aligns feature distributions, enabling accurate predictions without the need for labeled painting image data.
  \item Conduct comprehensive qualitative and quantitative experiments to assess the effectiveness of our proposed method on natural scene and painting image datasets. 
 
\end{itemize}

The rest of the paper is organized as follows: 

Section \ref{sec:related_works} provides an overview of previous works in the literature related to visual attention modeling, encompassing both saliency maps and scanpath prediction.
In Section \ref{sec:method}, we illustrate the components of our proposed model architecture, the training protocol, and the utilization of unsupervised domain adaptation for generalization. 
Section \ref{sec:results} presents the detailed results and evaluation of our model's performance and ablations of its components. Section \ref{sec:conclusion} ends the paper with key findings drawn from our study.

\section{Related works}
\label{sec:related_works}

In this section, we provide an extensive overview of related works on saliency prediction and scanpath detection, including both traditional and deep learning-based models.

\subsection{Saliency Prediction}

Saliency prediction has a long history, starting with the seminal work of \cite{KochUllman}, which was implemented by \cite{Itti} using a multi-scale model based on low-level features such as color, intensity, and orientation. Other models, like the one proposed by \cite{bruno2020multi}, build upon the extraction of key points for saliency prediction across multi-scale analysis.

The success of deep learning in the last decade led to the development of various deep architectural models for saliency prediction. For example, \cite{deepgaze1} used the MIT dataset \cite{MIT1003} to train a deep network, and they later improved their results in successive versions \cite{deepgaze2, deepgaze3}. Similarly, \cite{salicon} used a multi-scale architecture and a large dataset for training.

In \cite{unisal}, the authors introduced a unified model for saliency prediction (Unisal) applicable to static images and videos. They utilized Gaussian prior maps to model biases, leading to better predictions.

The field of saliency prediction has also been explored with graph-based methods. For instance, GBVS (Graph-Based Visual Saliency) \cite{GBVS}, proposed by \cite{Harel}, relies on graphs and Markov chains defined over different input maps. Meanwhile, \cite{infosal} proposed a model based on information maximization using the Shannon self-information measure.

Some works examined saliency prediction in the spectral domain, like \cite{spectral1} and \cite{spectral2}. In contrast, Bruno et al. \cite{bruno2020multi} used local keypoints in nearly perceptive uniform color spaces for saliency prediction.

The rise of deep learning paradigms positively impacted saliency prediction. For instance,   \cite{PanDSN} proposed a method for saliency prediction using deep and shallow neural networks. DeepGaze I \cite{deepgaze1} and its improved versions \cite{deepgaze2} utilized the MIT dataset and a VGG-19 network base to achieve accurate saliency map predictions.

Other deep learning approaches for saliency prediction rely on multi-level features from different levels of the network, as demonstrated in \cite{mlnet} and \cite{salicon}. Additionally, SalGan \cite{salgan} employed generative modeling and adversarial learning with an encoder-decoder architecture.

 \cite{sam} introduced a model that combined a convolutional neural network with recurrent Long Short-Term Memory (LSTM) to capture attention mechanisms. Unisal \cite{unisal} employed Gaussian prior maps to model biases for images and videos.

 \cite{tanssalnet} combined the use of CNNs with the Transformer architecture, this helps capture long-distance dependencies between image regions improving further the prediction.

Tliba et al. \cite{satsal} uses a U-net akin \cite{Unet} architecture and employs self-attention heads on the long skips connections. The model tries to merge low-level features extracted in the early stages and the high-level features from the latter. Using self-attention helps further refine the representational capability of the model.

\subsection{Scanpath Prediction}

Scanpath prediction is a relatively newer area of research but has gained significant interest in recent years.

The winner-take-all (WTA) module proposed by \cite{Itti} was one of the first methods to predict scanpaths. This approach extracted fixation points from saliency map peaks and inhibited the region surrounding these points.

Le Meur et al. \cite{lemeur}, introduced a stochastic approach   to generate scanpaths. In particular, they predicted saliency maps and modeled the probabilities of several biases, such as saccade amplitudes and orientations.

 \cite{G-Eymol} approached saliency maps as a 2D gravitational field affecting the trajectory of a mass representing the gaze. This gravitational field influences the speed and trajectory of the mass, which corresponds to the gaze movement.

Another approach was proposed by  \cite{shao}, who utilized high-level features from CNNs and Memory Bias, including short-term and long-term memory, for scanpath prediction.

 \cite{saltinet} presented Saltinet, a deep learning model that infers saccades from a static saliency volume generated by an encoder-decoder network. The inference is carried out using different sampling strategies, allowing the generation of scanpaths. Later, the same authors introduced PathGan \cite{pathgan}, which uses an LSTM network and conditional GAN architecture to predict scanpaths with increased accuracy.

DCSM (Deep Convolutional Saccadic Model) \cite{DCSM} predicts foveal saliency maps and temporal duration while modeling the Inhibition of Return (IoR), which accounts for the dynamic temporal aspect of scanpaths.

The SALYPATH model \cite{SALYPATH}, proposed an end-to-end model that simultaneously predicts scanpaths and saliency maps for an image. This approach exploits the correlation between the two tasks for improved performance. Later, the same approach was generalized for $360^\circ$ images \cite{Salypath360}.

The work of \cite{Tliba_2022_CVPR} employs a light architecture that extends central bias theory and employs Point-Net inspired 1D -convolutions to predict the adequate scanpaths. The authors used a self-supervised training method to extend the model on painting datasets and also introduced the preliminary version of the AVAtt (Art-Visual-Attention) dataset.  

The work of \cite{scanpath_domain}, used unsupervised domain adaptation \cite{MedDomainAdapt} \cite{DA} to transfer knowledge from a custom model trained on natural scene images to paintings. This work is an extension of that model. 

The authors also used Adversarial training in \cite{kerkouri2023interobserver} to improve the model performance and help capture the inter-observer patterns for consistent prediction.

In \cite{neurVA} the authors inspired by the biological model of the human visual system incorporated a foveated prediction model.

In \cite{scanpathnet}, the authors incorporate a dynamic priority map influenced by semantic content and fixation history. Using convolutional neural networks, convolutional long short-term memory networks, and mixture density networks, the model generates simulated human scanpaths.

In \cite{UMSS} The Unified Model of Saliency and Scanpaths (UMSS) is introduced as an approach that simultaneously predicts visual saliency and scanpaths in information visualizations.  UMSS analyzes gaze behavior for individual visualization elements and presents multi-duration element-level saliency maps. It then probabilistically samples scanpaths from these maps. 

\cite{scandmm} ScanDMM, a Deep Markov Model (DMM) architecture, for scanpath prediction in 360-degree images. It considers time dependency and incorporates a semantics-guided transition function to learn the nonlinear dynamics of attentional landscapes on omnidirectional images.

In conclusion, saliency prediction has a long history, with traditional and deep learning-based models achieving impressive results. Scanpath prediction, being a newer area of research, has seen significant progress in recent years, with approaches utilizing stochastic models, gravitational field analogies, high-level features, and dynamic temporal aspects for accurate predictions. The combination of saliency and scanpath prediction models allows a more comprehensive understanding of visual attention mechanisms and image understanding.

\section{Proposed method }
\label{sec:method}

\subsection{Architectural composition}
\label{subsec:arch-comp}

The proposed work introduces a novel framework for scanpath prediction and demonstrates the effectiveness of unsupervised domain adaptation on this task. The overall architecture of the proposed model is illustrated in Fig.\ref{figure:arch}. The key components and steps of the model are as follows:

1. \textbf{Feature Extractor} (Image Encoder): MobileNet is used as the feature extractor, responsible for capturing relevant patterns and features from the input images efficiently.

2. \textbf{Learnable Domain Prior Bias Maps}: Learnable domain prior bias maps are concatenated with the output of the feature extractor to help the model understand domain-specific characteristics during unsupervised domain adaptation.

3. \textbf{Merging Convolution}: The merged feature maps (from the feature extractor and bias maps) are processed through a series of convolutions to extract specific characteristics relevant to scanpath prediction.

4. \textbf{Soft-ArgMax}: The weighted feature maps are passed through the Soft-ArgMax function to generate continuous coordinates for predicted fixation points.

5. \textbf{Fixation Selector}: This component utilizes global pooling and fully connected layers to discretize the probability vector and generate a binary mask related to the relevant predicted fixation points.

6. \textbf{Stochastic Generation}: Randomness is introduced through noise sampling and modulation using a temperature parameter, adding stochasticity to the generation of scanpaths.

By combining these steps and components, the proposed model can predict scanpaths effectively and adapt to new domains through unsupervised domain adaptation. The stochastic generation adds randomness to the predictions, reflecting the subjective nature of human gaze patterns.

\subsubsection{Feature extractor}
\label{subsub:feature-extract}

Pixel-wise computations in gaze prediction involve analyzing and processing each pixel in the input images individually. That can be computationally intensive due to images' high-dimensional complexity and spatial correlation. To handle this complexity and reduce the computational burden, we use Convolutional Neural Networks (CNNs) to model and extract meaningful features from the input images.

CNNs \cite{neocognitron} \cite{cnn1}
are a class of neural networks well-suited for image-related tasks, as they can automatically learn hierarchical representations of the image data. By using CNNs, we can capture relevant patterns and features from the input images, which are crucial for accurate gaze prediction.

In our proposed method, we specifically use MobileNet \cite{mobileNet} as the feature extractor. MobileNet is a lightweight CNN architecture designed for mobile and embedded applications. It incorporates depth-wise separable and point-wise convolutions, significantly reducing the number of parameters and computations involved in the network. This makes the architecture more efficient and easier to train, especially with the limited amount of data available in our case.

By leveraging MobileNet as the feature extractor, our overall architecture remains lightweight while still being able to capture essential features from the input images. The choice of this network helps improve the efficiency and effectiveness of the gaze prediction model for the given dataset and computational resources.

\subsubsection{Priority maps and attention biases}
    \label{subsub:priors}

Saliency in images often exhibits a phenomenon known as center bias, where observers tend to focus their gazes more on the central regions of the image compared to the edges and corners. The so-called center bias phenomenon is commonly observed in eye-tracking experiments. It is believed to be a result of the human visual system's preference for attending to relevant semantic information, typically concentrated around an image's central regions. This stems from the anatomical distribution of photo-receptors on the retina \cite{photoreceptor_dist}.

To model the center bias phenomenon, a commonly used approach is to represent it using a 2-dimensional Gaussian function. The Gaussian function is a probability distribution that assigns higher probabilities to points nearer to the center and lower probabilities to points further away from the center. The mathematical form of the 2-dimensional Gaussian function is given by Equation \ref{eq:gaussian2d}:

\begin{equation}
f(x, y) = \frac{1}{2\pi \sigma_x \sigma_y} \exp \left( -\frac{(x-\mu_x)^2}{2\sigma_x^2} - \frac{(y-\mu_y)^2}{2\sigma_y^2} \right)
\label{eq:gaussian2d}
\end{equation}

In this equation, $x$ and $y$ are the coordinates of a point on the map, and $(\mu_x, \mu_y)$ and $(\sigma_x, \sigma_y)$ are the means and standard deviations of the distribution, respectively. The parameters $(\mu_x, \mu_y)$ control the location of the peak of the Gaussian, while $(\sigma_x, \sigma_y)$ determine the spread or width of the distribution.

In the context of our proposed model, we utilize these Gaussian functions to model the center bias in scanpath predictions. However, instead of using a single Gaussian function to represent the center bias, our model learns multiple maps, each following the same Gaussian law but with different means and standard deviations. These maps are referred to as "attention biases" or "priority maps."

By incorporating these attention biases into the model, we allow the feature extractor to focus on modeling the specific features of the stimuli images without being affected by biases related to the saliency task. In other words, the attention biases help the model to disregard the general center bias phenomenon that is present in eye-tracking data and instead concentrate on learning the saliency patterns that are specific to the given stimuli.

The integration of the attention biases is essential for achieving accurate and relevant scanpath predictions, as it allows the model to better understand and interpret the saliency patterns in the images. By learning the attention biases from the data, the model can effectively handle the center bias phenomenon and produce scanpath predictions that align more closely with human eye movement behavior.

\subsubsection{Scanpath prediction}
\label{subsub:scanpath-gen}

The scanpath prediction decoding process is composed of several components, namely the merging module, fixation selection network, and the Soft-ArgMax function.

The merging module is a network that consists of 8 ($3 \times 3$) convolutional layers, each activated by a Rectified Linear Unit (ReLU) function. This module takes as input the feature maps extracted by the MobileNet feature extractor, which are concatenated with the learnable prior maps (attention biases) described earlier. The merging network combines the stimuli-specific features extracted by MobileNet with the prior maps while gradually reducing the number of feature maps to 20.

An activation probabilities vector is generated by the channel selection network, which is used to weigh the resulting feature maps from the merging module. The channel selection network is responsible for predicting fixation point coordinates using the Soft-ArgMax Function (SAM) \cite{sam}, similar to the approach in \cite{SALYPATH}.

The Soft-ArgMax function predicts the coordinates of fixation points from the model's output feature maps. It is mathematically described as follows:

\begin{equation}
    SAM(x) = \sum_{i=0}^{W} \sum_{j=0}^{H} \frac{e^{\beta x_{i,j}}}{\sum_{i^\prime = 0}^{W} \sum_{j^\prime = 0}^{H} e^{\beta x_{i^\prime, j^\prime}}} \left(\frac{i}{W}, \frac{j}{H}\right)^T
    \label{eq:SAM}
\end{equation}

In this equation, $x$ is the input feature map, and $(i, j)$ and $(i^\prime, j^\prime)$ iterate over pixel coordinates. $W$ and $H$ represent the height and width of the feature map, respectively. The parameter $\beta$ adjusts the distribution of the softmax output.

The fixation selector is another crucial component that generates a binary mask vector to select suitable fixation points to be part of the predicted scanpath. This allows for generating variable-length scanpaths. The configuration of the fixation selector involves passing the feature maps from the merging module through Global Max Pooling and Average Pooling layers in parallel after flattening. The resulting vectors are used as inputs to a shallow  Multi-Layer Perceptron (MLP) network. The outputs of the two branches of the MLP are concatenated and passed to a third MLP Layer, which is activated by a Sigmoid function in the last layer to provide a vector with values in the continuous range $[0, 1]$. This vector is then dynamically binarized based on its mean value to create the binary mask vector,  as shown in the following equation:  

\begin{equation}
    mask_b= v > mean(v)  
    \label{eq:b_mask}
\end{equation}
with
\begin{multline}
    v = MLP_3(cat( MLP_1(GMaxPool( flat(d))) , \\ MLP_2(GAvgPool(flat(d)))))      
    \label{eq:p_mask}
\end{multline}

where $mask_b$ is the resulting binary mask vector. $MLP_1, MLP_2$  and  $MLP_3$ are the shallow MLP networks, $cat$ is a vector concatenation, $GMaxPool$  is the global max pooling operation,  $GAvgPool$  is the global average pooling operation, $flat$ is a flattening function and $d$ is the output feature maps obtained from the Merging layers.

{
\color{black}
Because of the non-differentiability of the binarization function, the network's computational graph is cut stopping the flow of gradients. To address this issue, the vector obtained from the MLP is multiplied by the feature maps. This ensures that the binarization function's non-differentiability is handled effectively. In other words, since the binarization function is non-differentiable, we need to find an alternative way to allow the flow of gradients. Hence, we multiply the results of the fixation selection network with the input of the SAM function. This step enables us to retain the necessary gradient information and ensure smooth training and optimization.
}

To address the non-differentiability of the binarization function, the network's computational graph is cut, and the vector obtained from the MLP is multiplied by the feature maps. The Sigmoid activation in the last layer of the MLP ensures that the operations remain within the same range as the network's output, avoiding the exploding gradient problem during back-propagation. This fixation selector network plays a crucial role in generating accurate scanpath predictions, as it effectively selects the relevant fixation points based on the feature maps, producing a scanpath that closely resembles human eye movement behavior.

\subsection{Training and Loss}
\label{sec: Training}
    
The training process involved  9000 images from the Salicon dataset \cite{salicon}, while 1000 additional images were set aside for validation purposes.

To train the model, we employed the following loss function:

\begin{equation}
    L(y, \hat{y}) = BCE(y, \hat{y}) + 0.001 \times \sqrt{\text{len}(y)^2 - \text{len}(\hat{y})^2} 
    \label{eq:loss}
\end{equation}

In this equation, $y$ represents the predicted scanpath vector, $\hat{y}$ is the ground truth scanpath vector, and $BCE$ denotes the Binary Cross-Entropy function.

The loss function is designed to capture two aspects of scanpath prediction. Firstly, it penalizes the model for misalignments with the ground truth by using the Binary Cross-Entropy loss. Secondly, it introduces a regularization term based on the difference in length between the predicted scanpath and the ground truth scanpath.

The Adam optimizer is utilized for training, with a learning rate of $5 \times 10^{-5}$. We trained the model for 70 epochs and initialized all the weights using the Xavier initializer. This training setup allowed the model to learn and adapt its parameters effectively to predict accurate and representative scanpaths. The validation process during training ensured that the model's performance could be monitored and optimized throughout the training process.

\begin{figure*}
    \centering
    \includegraphics[width= 0.9\linewidth]{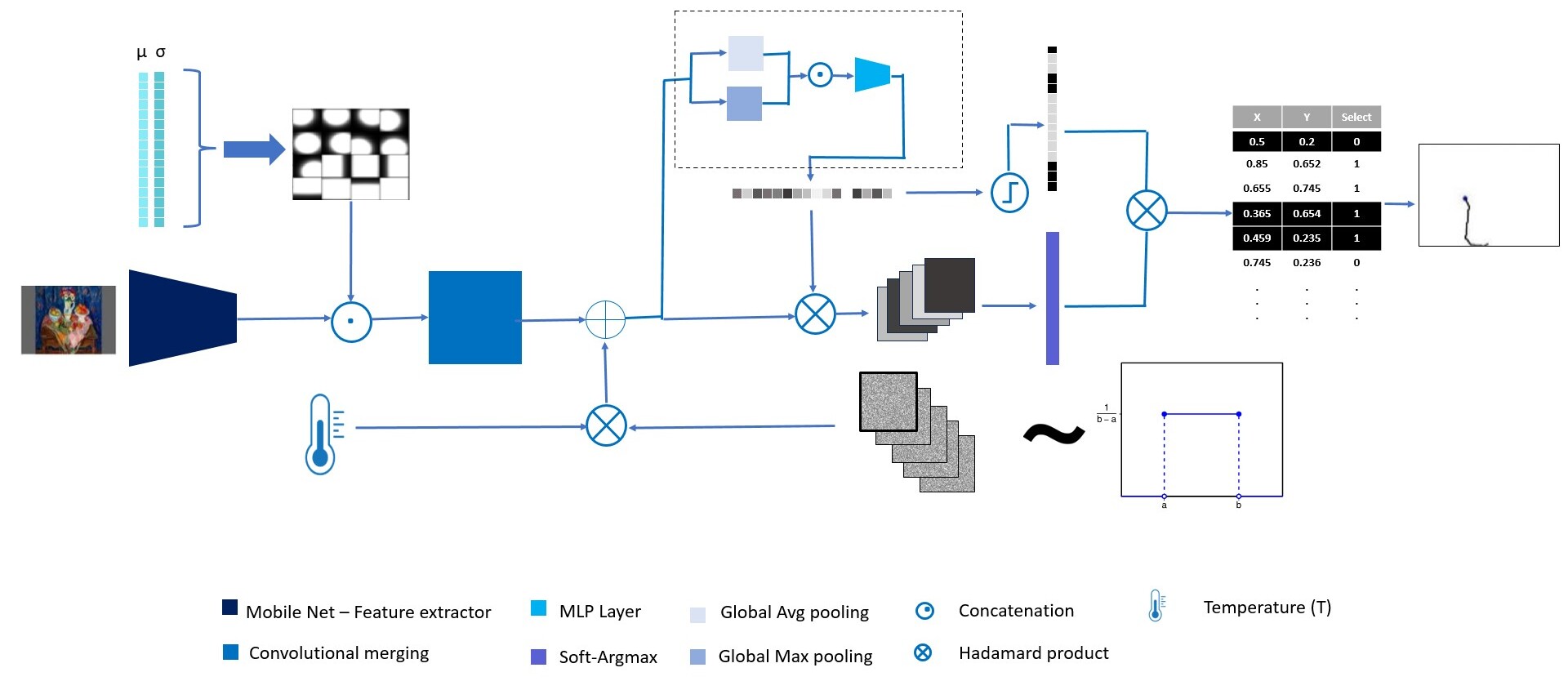}
    \caption{\label{figure:arch_stoch}General architecture of the proposed method.}
\end{figure*}

\subsection{Domain Adaptation}
\label{sub:domain-adaptation}

The initial training of our model on natural scene images yielded promising results, as shown in Section \ref{sec:results}. However, when we tested the same model on images from other source domains, such as paintings, the performance was not as satisfactory. To fully harness the capabilities of our model, we need to adapt it to new source domains.

Domain adaptation involves training a model on a source domain (e.g., natural scene images) and then adapting it to perform well on a target domain (e.g., paintings). The key challenge is to minimize the discrepancy between the distributions of the two domains so that the model can generalize effectively to the target domain. For this purpose, we adopt an adaptation method proposed in \cite{DA}.

To enable domain adaptation, we want our feature extractor to be capable of identifying shared and useful information between the two domains while discarding domain-specific features and noises that are not relevant for the prediction task. To achieve this, we introduce a small branch to the network that classifies images as belonging to either the natural scene domain or the painting domain. This branch is added in a pseudo-labeled manner, meaning that we do not have explicit labels for the domain, but we treat the source and target domains as separate classes.

To encourage the feature extractor to find a unified representation space between the two domains, we use a gradient reversal layer (GRL) at the start of the domain classifier branch. The GRL reverses the sign of the gradient flow during back-propagation. During forward propagation, it has no effect (Eq. \ref{eq:GRL_forward}), but during back-propagation, it multiplies the gradients by -1 (Eq. \ref{eq:GRL_backward}).

\begin{equation}
    GRL(x) = x   
    \label{eq:GRL_forward}
\end{equation}

\begin{equation}
    GRL(\frac{\partial L}{\partial x}) = - \frac{\partial L}{\partial x}   
    \label{eq:GRL_backward}
\end{equation}

Where $x$ is the input of the layer, and $\frac{\partial L}{\partial x}$ represents the gradient of the domain loss $L_d$ when backpropagating through the network.

As shown in Fig.\ref{figure:Da-arch}, while the domain classifier learns to discriminate between images from the two domains, the reversal of the gradient sign in the GRL encourages the feature extractor to find a shared representation space, minimizing the distance between the distributions of the two domains.

In essence, the domain adaptation process forces the feature extractor to focus on the mutual characteristics of the two domains while ignoring domain-specific variations, effectively aligning the distributions of the two domains.

The unified representation space can be conceptualized as the union of the two distributions (natural scene images and paintings) minus the specific noise distributions of each domain:

\begin{equation}
    \mathcal{D} = \{ \mathcal{D}_p \cup \mathcal{D}_n \} \setminus \{ \mathcal{N}_p \cup \mathcal{N}_n \}  
    \label{eq:dists}
\end{equation}

where:\\
- $\mathcal{D}_n$ and $\mathcal{D}_p$ are the distributions of natural scene images and paintings, respectively.\\
- $\mathcal{N}_n$ and $\mathcal{N}_p$ are the noise distributions specific to the natural scene and painting domains.

For unsupervised domain adaptation training, we used a mix of 2000 Salicon images \cite{salicon} and 2000 unlabeled paintings images scraped from the internet. This allows the model to adapt to the characteristics of the painting domain and perform well on both natural scene and painting images.

\subsection{Stochastic Inference}
\label{subsec:Stochastic}

Visual attention is a complex and subjective phenomenon influenced by various physiological, psychological, cognitive, environmental, and situational factors. As a result, we observe significant variations in gaze patterns and scanpaths taken by multiple subjects or even the same individual in different situations or over time. These variations cannot be fully controlled or accounted for during eye-tracking experiments, indicating that the prediction of scanpaths is a stochastic generative task.

To introduce randomness and stochasticity into our model's predictions, we modify the network architecture as shown in Fig. \ref{figure:arch_stoch}. We introduce a new element in the form of a random latent vector $L \sim U(\text{to be filled})$, where $L \in \mathbb{R}^{w \times h \times c}$. This random vector captures the stochastic nature of the scanpath generation process. We then modulate the randomness using a variable $T$, which we refer to as the temperature. The temperature parameter controls the level of stochasticity in the generated scanpaths.

We can model the generated scanpath as follows:

\begin{equation}
    y = \text{decoder}(\text{encoder}(x) + L \times T)
\end{equation}

Where:

$x$ is the input image.
$\text{encoder}$ is the encoder part of the model that transforms the input image into a latent representation (i.e. Feature extractor).
$\text{decoder}$ is the decoder part of the model that generates the predicted scanpath from the latent representation and the modulated randomness $L \times T$.
By introducing the random latent vector and modulating its effect with the temperature, we can incorporate stochasticity into the scanpath generation process. This allows the model to generate diverse and varied scanpaths for the same input image, reflecting the inherent variability observed in human gaze behavior. The temperature parameter controls the balance between deterministic and stochastic behavior in the model's predictions, providing flexibility in generating scanpaths with different levels of randomness.

\section{Experimental Protocol}
\label{sec:results}

    \subsection{Datasets}
    \label{subsec:datasets}
In this study, we leverage two widely recognized natural scene datasets, namely Salicon \cite{salicon} and MIT1003 \cite{MIT1003}. Additionally, we utilize two painting datasets, namely the Le Meur painting dataset \cite{paint_lemeur} and the Art-Visual-Attention (AVAtt) dataset \cite{Tliba_2022_CVPR} \cite{Avatt}. These datasets serve as the basis for our experimental evaluation and enable us to assess the model's performance across diverse visual domains, encompassing both natural scenes and paintings.

\subsubsection{Salicon Dataset}
\label{subsub:salicon}

The \textbf{Salicon} dataset is a large-scale and valuable resource extracted from the MS-COCO (Common Objects in Context) dataset \cite{Coco}. It comprises a diverse collection of images, each accompanied by its corresponding saliency map and a set of recorded scanpaths ($\approx 50$). This dataset serves as an essential benchmark for evaluating and advancing visual attention modeling methods.

In our experiments, we partitioned the Salicon dataset into three subsets: a training set containing 9,000 images, a validation set with 1,000 images, and a test set comprising 5,000 images. The training and validation sets facilitated the training process of our proposed model. While, the test set played a critical role in assessing the model's performance and comparing it with other state-of-the-art models in the field.

Throughout our work, we extensively evaluated our model by testing on the approximately  250,000  available scanpaths from the Salicon dataset\cite{salicon}. This substantial number of scanpaths ensures the soundness and robustness of our empirical results, as it allows us to thoroughly examine various aspects of our model's performance. The extensive evaluation process, combined with the large and diverse dataset, contributes to the validity and reliability of our findings, providing a comprehensive understanding of the effectiveness and capabilities of our proposed model.

\subsubsection{MIT1003 Dataset}
\label{subsub:mit1003}

The \textbf{MIT1003} dataset is widely recognized as one of the most prominent datasets for static image saliency. Often used alongside the \textbf{MIT300} dataset \cite{MIT300}, MIT1003 consists of 1,003 natural scene images that depict various objects, individuals, and scenes under diverse conditions. Each image in the dataset is accompanied by its corresponding saliency map and a collection of scanpaths recorded from 15 different observers, resulting in a total of 15,045 scanpaths available for testing.

In our study, we employ the entire MIT1003 dataset primarily to assess the generalization ability of our model across different datasets, specifically for cross-dataset evaluation. To achieve this, we refrain from fine-tuning our model on this dataset, focusing solely on evaluating its performance on previously unseen data. This approach allows us to measure the model's ability to generalize its predictions beyond the training data, gauging its adaptability to novel visual scenes and diverse conditions encountered in the MIT1003 dataset.

By conducting our evaluation on the MIT1003 dataset without any fine-tuning, we can confidently analyze the robustness and effectiveness of our model's generalization capabilities, providing insights into its capacity to make accurate predictions on previously unseen data. The extensive collection of scanpaths and diverse nature of the images in this dataset contribute to the comprehensive evaluation of our model's performance and demonstrate its potential for real-world applications.

\subsubsection{Le Meur painting Dataset}
\label{subsub:paintLemeur}

The \textbf{Le Meur paintings} dataset \cite{paint_lemeur} is a valuable resource specifically designed for saliency modeling in the domain of paintings. It consists of 150 images of paintings, each accompanied by its corresponding saliency map. This dataset serves as a critical evaluation benchmark to assess our model's capacity to transfer knowledge learned from natural scene images to the distinct domain of paintings. Furthermore, it allows us to test the effectiveness of our domain adaptation techniques on this unique dataset.

While the \textbf{Le Meur paintings} dataset does not provide observers' scanpaths, it remains one of the few publicly available visual attention dataset for paintings. Despite the absence of scanpaths, this dataset serves as an essential resource for evaluating the generalization ability of our model in the painting domain. By applying our domain adaptation techniques, we can explore how well the model adapts to the unique characteristics of paintings and whether it can accurately predict saliency in this specialized domain.

The \textbf{Le Meur paintings} dataset plays a crucial role in broadening the scope of our study beyond natural scene images and demonstrates the versatility of our model in handling diverse visual domains. By leveraging this dataset, we gain valuable insights into the model's capability to generalize its knowledge and effectively predict saliency in the distinct context of paintings, contributing to a deeper understanding of visual attention in various artistic and visual content scenarios.

\subsubsection{Art-Visual-Attetnion Dataset}
\label{subsub:paintLemeur}

The Art-Visual-Attention dataset (AVAtt) \cite{Tliba_2022_CVPR} \cite{Avatt} comprises 164 images, each accompanied by its corresponding saliency map and 15 observer scanpaths for every image. As the sole available dataset with individual scanpath annotations for each image, AVAtt stands as a valuable resource for studying visual attention in the context of artwork.

Although the AVAtt dataset is currently in the process of being enriched and has not been publicly shared yet, it plays a pivotal role in our study. We utilize this dataset to assess the generalizability of our model on new painting images, specifically evaluating how well our model can adapt to diverse artistic styles and art movements. The dataset provides a diverse range of images from 16 different art movements, depicting a wide variety of subjects.

By testing our model on the AVAtt dataset, we gain crucial insights into its ability to generalize its knowledge from training on other datasets to the unique characteristics and styles present in artwork. The inclusion of individual scanpath annotations for each image further enhances the dataset's value, enabling us to analyze the model's performance on a more detailed level and understand how it predicts visual attention in the context of paintings with various artistic influences.

Overall, the AVAtt dataset serves as a key resource for exploring the generalizability and adaptability of our model to new painting images, providing valuable information for real-world applications in art analysis, visual content design, and other domains where understanding human visual attention in artwork is crucial.

    \subsection{ Evaluation metrics}
    \label{sub:metrics}

In our evaluation, we utilized the same set of images and partitions as described in Section \ref{subsec:datasets} to test both our model and the compared methods. To assess the performance of each model, we compared their results to the ground truth using three evaluation metrics: $MultiMatch$ \cite{Multimatch}, $NSS$ \cite{NSS}, and $Congruency$ \cite{congruency}.

\subsubsection{Multimatch metric}
\label{subsub:multimatch}

\textbf{MultiMatch} (Jarodzka) was introduced by Jarodzka et al. \cite{Multimatch}. This metric measures the similarity between scanpaths as multi-dimensional vectors and compares them based on five different characteristics:

1. \textbf{Shape}: It quantifies the similarity in shape between the scanpaths represented as vectors, denoted as $\overline{u} - \overline{v}$. This calculation captures the structural variation between the predicted and ground truth scanpaths, allowing for a comprehensive comparison of their spatial configurations.

2. \textbf{Direction}: It assesses the difference in angle between saccades in the scanpaths. This evaluation ensures that the predicted saccadic directional changes align with the distribution observed in human gaze behavior, verifying the model's ability to produce realistic and natural gaze shifts.

3. \textbf{Length}: It evaluates the discrepancy in length between saccades. This assessment quantifies the variation in the extent of gaze shifts between the model-predicted and observed scanpaths, providing valuable insights into the model's accuracy in reproducing the spatial characteristics of human eye movements.

4. \textbf{Position}: It computes the dissimilarity in positions between the fixation locations in the scanpaths. This calculation allows for a thorough evaluation of how well the model's predicted gaze points align with the actual fixation positions made by human observers. By quantifying the positional differences, this characteristic provides a measure of the model's accuracy in capturing the spatial distribution of visual attention across the observed scanpaths.

5. \textbf{Duration}: It assesses the discrepancy in saccade durations between the model-predicted scanpaths and the observed scanpaths. This evaluation allows for a comprehensive analysis of the model's ability to capture temporal aspects of visual attention and the timing of gaze shifts observed in human eye movements. However, since our model does not predict the time dimension, we excluded this component from our evaluation. Thus, we focus solely on the other characteristics of the \textbf{MultiMatch} metric to assess the model's performance in replicating the spatial aspects of human gaze transitions during visual exploration.

To make scanpath comparisons feasible, the MultiMatch metric involves three crucial steps. First, the scanpaths are simplified by condensing small, locally contained saccades and fixations. This critical stage ensures that minor differences do not unduly influence the overall similarity result. Second, the simplified scanpaths are temporally aligned, establishing correspondence between pairs of saccades for subsequent comparison. The Dijkstra algorithm effectively identifies the optimal alignment by minimizing the costs based on differences between potential saccade pairings. Finally, the aligned scanpaths undergo detailed comparison, considering five dimensions, including saccade amplitudes, directions, and fixation durations as mentioned above. 

Each characteristic produces a value between 0 and 1, where a lower value indicates greater similarity between the scanpaths.

To obtain an overall score representing the complete similarity, an arithmetic mean is calculated for the first four characteristics (i.e. Shape, Direction, Length, and Position), excluding Duration since our models do not predict fixation timestamps. The overall score, denoted as \textit{MM Score}, ranges between 0 and 1, with 1 indicating high similarity and 0 indicating low similarity.

The MultiMatch metric offers a comprehensive evaluation of scanpath similarity, allowing us to assess the performance of our models in capturing the essential characteristics of human visual attention patterns and comparing them to ground truth scanpaths.

\subsubsection{NSS Metric}
\label{subsub:NSS-metric}

\textbf{NSS} (Normalized Scanpath Saliency) \cite{NSS} is a hybrid evaluation metric used to compare saliency maps and predicted scanpaths. The process involves constructing a binary fixation map based on the predicted scanpath. This binary map is then multiplied pixel-wise with a normalized saliency map, which has a mean of zero and a standard deviation of one. The arithmetic mean of the resulting multiplication provides the NSS score, representing the mean saliency value of the fixation points in the scanpath.

The NSS score is given by the following equation:
\begin{equation}
    NSS = \frac{1}{N} \sum (P \odot Q_b)
    \label{eq:NSS}
\end{equation}

where:
\begin{itemize}
    
    \item  \textbf{N} is the total number of fixation points in the scanpath.
    
    \item \textbf{Q\textsubscript{b}} is the binary fixation map derived from the scanpath.
    
    \item \textbf{S} is the original saliency map.
    
    \item \textbf{P} is the normalized saliency map, calculated as \(\frac{S - \mu(S)}{\sigma(S)}\).

    \item $\mu(S)$ and $\sigma(S)$ represent the mean and standard deviation of the saliency map, respectively.
\end{itemize}

The NSS score provides a measure of how well the model-predicted scanpath corresponds to the general saliency distribution in the saliency map. A higher NSS score indicates better alignment between the predicted fixation points and the regions of higher saliency in the saliency map, demonstrating the model's accuracy in reproducing the saliency-driven attention patterns observed in human observers.

\subsubsection{Congruency Metric}
\label{subsub:Congruency-metric}

The \textbf{Congruency} metric is a hybrid evaluation method that measures the agreement between the predicted scanpaths and the ground truth by evaluating their spatial overlap and similarity. Higher \textbf{Congruency} scores indicate a stronger alignment with the observed gaze patterns, indicating that the model accurately reproduces human attentional focus.

To calculate the \textbf{Congruency} score, the following steps are performed:

\begin{equation}
    Congruency = \frac{ \sum (P_{otsu} \odot Q_b)}{\sum Q_b}
    \label{eq:congruency}
\end{equation}

Where
\begin{itemize}

    \item \textbf{Q\textsubscript{b}} is the binary fixation map derived from the scanpath or scanpaths.
    
    \item \textbf{S} is the original saliency map.
    
    \item \textbf{P} is the normalized saliency map, calculated as \(\frac{S - \mu(S)}{\sigma(S)}\).

    \item $\boldsymbol{P_{otsu}}$ 
    is the binarized saliency map, calculated by applying the Otsu  thresholding algorithm \cite{otsu}.

\end{itemize}

The \textbf{Congruency} score provides insights into how well the model-predicted fixations align with the regions of high saliency in the stimulus. A higher \textbf{Congruency} score indicates better congruence between the model's fixation locations and the salient regions identified by human observers. This metric is particularly useful for evaluating how well the model captures the attentional focus on visually salient areas in the stimulus, reflecting the model's ability to reproduce human-like attention patterns.

\begin{figure}
    \centering
    \includegraphics[width=\linewidth]{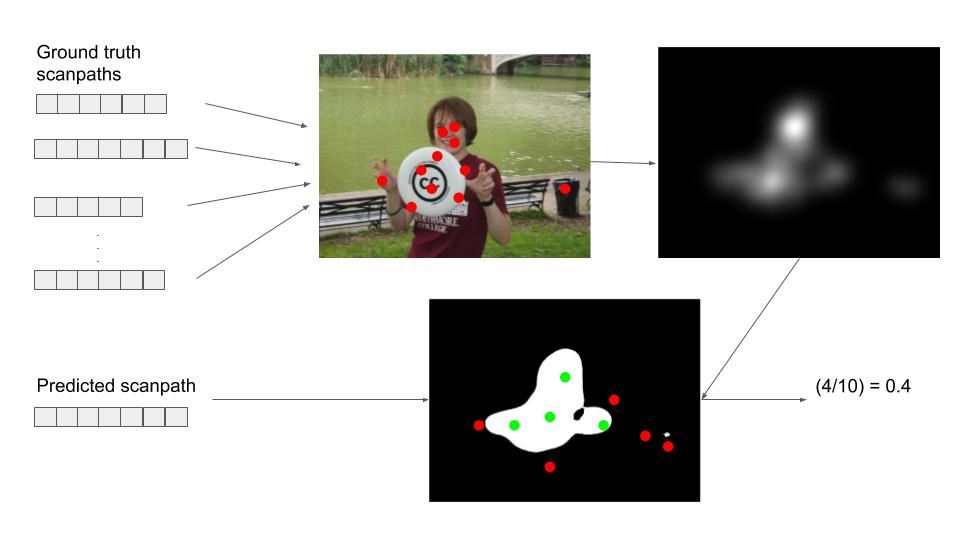}
    \caption{\label{figure:congruency} Calculating congruency of predicted scanpath.}
\end{figure}

In Fig. \ref{figure:congruency}, we summarize the principal steps of the \textbf{Congruency} metric. A fixation map is first constructed from the ground truth scanpaths and processed with a Gaussian kernel convolution to generate a saliency map for the image. This saliency map is then dynamically thresholded using Otsu's method \cite{otsu} to separate salient and non-salient regions. The \textbf{Congruency} is calculated as the ratio of fixation points within the salient region of the binary map to the total number of fixation points in the scanpath.

Unlike NSS, the \textbf{Congruency} metric does not take into account the saliency value of each fixation point, treating all salient points equally in the evaluation process.

\subsection{Experimental protocol overview}
\label{subsub:protocol}

To validate our model we use the dataset in 
 section \ref{subsec:datasets} and metrics described in section \ref{sub:metrics}. 

In order to demonstrate the effectiveness of our discriminative model for natural scene images ( Section \ref{subsec:detr-results}), we employ the 5000 images from the Salicon dataset \cite{salicon} with the corresponding $\approx 250,000$ corresponding scanpaths. We also used the 1003 images from  MIT1003 \cite{MIT1003} with the corresponding $15045$ scanpaths.

This would bring the theoretical number of pairs to compare for the Salicon dataset to ($ \approx 3,750,000$) comparisons. Because of the limitation of computational resources, and the computational heaviness of Multimatch metric \cite{Multimatch},  which does not have any public implementations benefiting from hardware acceleration. 

We chose to evaluate our generative model only on the MIT1003 dataset \cite{MIT1003} in section \ref{subsec:gener-results}. This brings ($225,675$) comparisons for evaluation. This decision was also instructed by the fact that MIT1003 is a true eye-tracking dataset while Salicon uses mouse movements.

We evaluated the generative model (Section \ref{subsec:gener-results}) using five temperature values chosen qualitatively by the authors; we used  normal and uniform distributions to generate the scanpath noise. 

In section \ref{sub:ablation}, we conduct an ablative study of the components used in our determinative architecture. 

We test the effectiveness of our domain adaptation method on the determinative method. Then evaluate the generative architecture that benefits from the domain adaptation.  

Finally, in section \ref{subsec:qualitative}, we offer some qualitative results for our method.

\subsection{Quantitative evaluation of the determinative network}
\label{subsec:detr-results}

\begin{table*}[h!]
\begin{center}
\begin{tabular}{ c c c c c c c c }
\hline
\textbf{Model} & \textbf{Shape } & \textbf{Direction} & \textbf{Length} & \textbf{Position} & \textbf{MM Score} & \textbf{NSS} & \textbf{Congruency}\\ 
\hline
 PathGAN \cite{pathgan} & 0.9608 & 0.5698 & 0.9530 & 0.8172 & 0.8252  & -0.2904  &  0.0825 \\ 
 \hline
 Le Meur \cite{lemeur} & 0.9505 & 0.6231 & 0.9488 & 0.8605 & 0.8457 &  0.8780 &   0.4784   \\ 
 \hline
 G-Eymol \cite{G-Eymol} & 0.9338 & 0.6271 & 0.9521 & 0.8967 &  0,8524 &  0.8727 & 0.3449 \\
 \hline
 SALYPATH \cite{SALYPATH}  & 0.9659  & \textbf{0.6275} & 0.9521 & 0.8965  &   0,8605  & 0.3472 &  0.4572     \\ 
 \hline
 Our Model  & \textbf{0.9702} & 0.6173 & \textbf{0.9587} & \textbf{0.8968} &  \textbf{0,8607} &  \textbf{1.0140} &   \textbf{0.5170} \\
 
 \hline
 
\end{tabular}

\caption{ \label{tab:MM-salicon}Results of scanpath prediction on Salicon dataset}
\end{center}
\end{table*}

\begin{table*}[h!]
\begin{center}
\scalebox{0.9}{
\begin{tabular}{ c  c  c  c  c  c  c c}
\hline
\textbf{Model} & \textbf{Shape } & \textbf{Direction} & \textbf{Length} & \textbf{Position} & \textbf{MM Score} & \textbf{NSS} & \textbf{Congruency}\\ 
\hline
 PathGan \cite{pathgan} & 0.9237  & 0.5630   & 0.8929   &  0.8124    &   0.7561  & -0.2750  &  0.0209        \\ 
 \hline
 DCSM (VGG) \cite{DCSM}  & 0.8720 & 0.6420 & 0.8730 & 0.8160 & 0,8007 & - & - \\ 
 \hline
 DCSM (ResNet) \cite{DCSM} & 0.8780 & 0.5890 & 0.8580 & 0.8220 &  0,7868 & - & -\\ 
 \hline
 Le Meur \cite{lemeur} & 0.9241  &  0.6378  & \textbf{0.9171}  &  0.7749 &  0,8135  & 0.8508   &  0.1974    \\ 
 \hline
 G-Eymol \cite{G-Eymol} & 0.8885 & 0.5954 & 0.8580 & 0.7800 &  0,7805 & 0.8700 & 0.1105\\
 
 \hline
 SALYPATH \cite{SALYPATH} & 0.9363  & \textbf{0.6507} & 0.9046 & 0.7983  &   0,8225 &  0.1595  &   0.0916    \\ 
 \hline
 Our model & \textbf{0.9392} & 0.6152  & 0.9100 & \textbf{0.8537}   &  \textbf{0.82952} &  \textbf{0.8888} &   \textbf{0.2114}   \\
\hline

\end{tabular}}
\caption{\label{tab:mit1003_results}Results of scanpath prediction on MIT1003.}
\end{center}
\end{table*}

Table \ref{tab:MM-salicon} presents the results of scanpath prediction on the Salicon dataset. The highest scores are highlighted in bold. Our model achieves the best results in the $Shape$, $Length$, and $Position$ components while demonstrating acceptable results for the $Direction$ component. It slightly surpasses the runner-up on the \textit{MM Score}, representing the four components' mean score.

Our model's performance on the $MultiMatch$ metrics is comparable to state-of-the-art models, even slightly outperforming SALYPATH in some criteria. Regarding the $NSS$ metrics, our model achieves the best results, surpassing Le Meur and G-Eymol, which use saliency maps for their scanpath generation. However, our model is surpassed by Le Meur and SALYPATH on the $Congruency$ metric. This can be attributed to the fact that Le Meur heavily relies on saliency maps for scanpath generation, while SALYPATH learned saliency intermediate features from the saliency prediction branch.

In summary, our model shows promising results on the Salicon dataset, particularly excelling in certain components of the $MultiMatch$ metric and obtaining the best performance in the $NSS$ metric, reflecting its ability to accurately capture visual attention patterns. However, there is room for improvement in the $Congruency$ metric, where other models with a stronger reliance on saliency maps outperform our approach.

Table \ref{tab:mit1003_results} displays the results of scanpath prediction on the MIT1003 dataset. Our model achieves the highest scores in the $Shape$ and $Position$ criteria, indicating its strong performance in accurately predicting the shape and position of scanpaths compared to other models. While our model's performance on the $Direction$ and $Length$ characteristics is slightly lower, it remains competitive when compared to state-of-the-art methods. Notably, our model surpasses other models in terms of the mean score, denoted as $MM Score$.

In terms of the $NSS$ and $Congruency$ scores, our model exhibits a slight decrease, but still outperforms the comparison models. The $NSS$ score reflects our model's ability to align with salient regions, while the $Congruency$ score indicates the agreement between predicted and observed scanpaths.

Overall, our model demonstrates promising results on the MIT1003 dataset, with competitive performance in various scanpath characteristics. It excels in accurately capturing the shape and position of scanpaths, making it a strong contender among state-of-the-art methods. Additionally, it maintains a good level of alignment with salient regions and gaze patterns, showcasing its ability to reproduce human-like attentional focus in static images.

\subsection{Quantitative Components Ablation}
\label{sub:ablation}

\begin{table*}[h!]
\begin{center}
\begin{tabular}{ c c c c c c c c }
\hline
\textbf{Model} & \textbf{Shape } & \textbf{Direction} & \textbf{Length} & \textbf{Position} & \textbf{MM Score} & \textbf{NSS} & \textbf{Congruency}\\ 
\hline
 Our model & \textbf{0.9392} & \textbf{0.6152}  & \textbf{0.9100} & \textbf{0.8537}   &  \textbf{0.82952} &  \textbf{0.8888} &   \textbf{0.2114}   \\
\hline
Our model (prior) &   0.9372   &   0.5850  &   0.9021   &   0.8519&    0.8188 &   0.8747   &   0.2007   \\
\hline
Our model (select) & 0.9216  & 0.5831 & 0.9087 & 0.8111 &  0.8061 & 0.9563  &   0.2326   \\
\hline

\end{tabular}
\end{center}

\caption{\label{tab:mit1003_ablation}Results of scanpath prediction on MIT1003 ablation.}
\end{table*}

In this section, we present the results of the quantitative components ablation study, which aims to analyze the impact of different components in our model on scanpath prediction performance. Table \ref{tab:mit1003_ablation} summarizes the results of the ablation study on the MIT1003 dataset.

\subsubsection{Prior maps ablation}
\label{subsub:prior-ablation}
we investigate the impact of prior maps and attention biases on the performance of our model. We compare the results of the full model (which includes prior maps and attention biases) with a variant of the model where we exclude these components. The results are shown in the row labeled "Our model (prior)" in Table \ref{tab:mit1003_ablation}.

By removing the prior maps and attention biases, the model's performance slightly decreases across all metrics compared to the full model. This suggests that prior maps and attention biases contribute positively to the model's ability to generate accurate scanpaths. These components help guide the model's attention toward important regions in the input image, leading to better predictions of human-like gaze patterns.

\subsubsection{Fixations selector ablation}
\label{subsub:prior-ablation}

\begin{figure*}
     \centering
     \begin{subfigure}[b]{0.3\textwidth}
         \centering        \includegraphics[width=\textwidth]{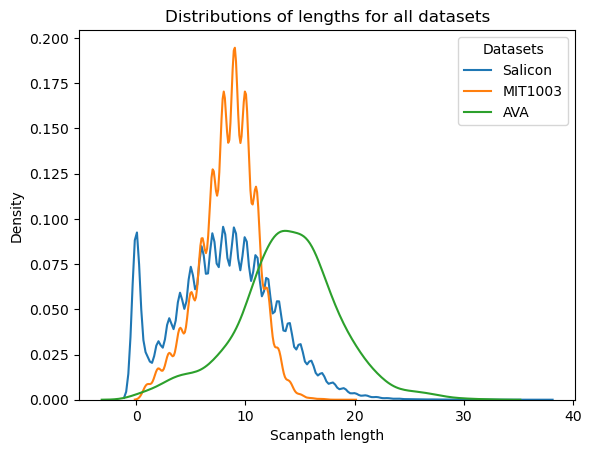}
         \caption{Length distributions of all datasets}
         \label{fig:length-dist-all}
     \end{subfigure}
     \begin{subfigure}[b]{0.3\textwidth}
         \centering        \includegraphics[width=\textwidth]{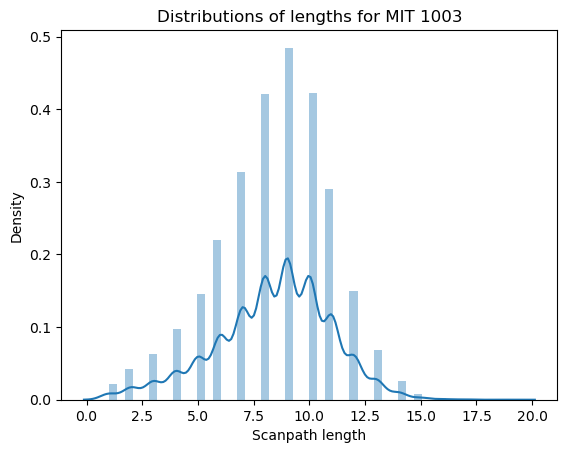}
         \caption{Original MIT1003}
         \label{fig:length-dist-org}
     \end{subfigure}
     \begin{subfigure}[b]{0.3\textwidth}
         \centering
        \includegraphics[width=\textwidth]{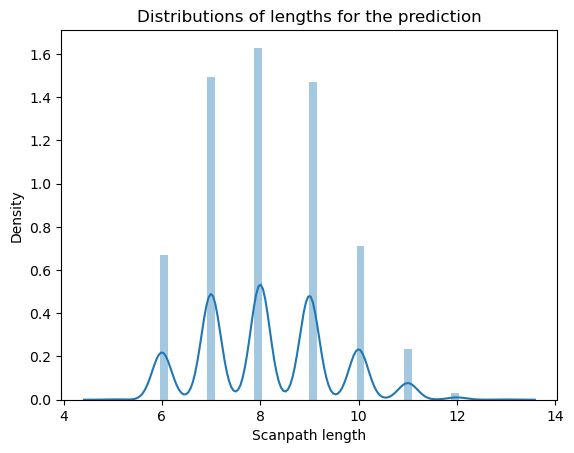}
         \caption{ Predicted MIT1003}
         \label{fig:length-dist-pred}
     \end{subfigure}
        \caption{Distributions of lengths of scanpaths on MIT1003.}
        \label{fig: lengths-dist  }
\end{figure*}

To analyze the importance of the selective attention mechanism, we compare the performance of our full model with the model where we exclude the fixations selector. The results are shown in the row labeled "Our model (select)." Without the selective attention mechanism, the performance decreases significantly in $MM Score$,  suggesting that the fixations selector plays a crucial role in improving the model's ability to capture human-like gaze patterns.

On the other hand, we notice that the $NSS$ and $Congruency$ values have increased. This is due to the simple nature of the component. Excluding this module would actually increase the number of fixations generated and thus increases these measures which are directly correlated to the number of fixations and their position.

Fig. \ref{fig: lengths-dist  } illustrates the distribution of scanpath lengths. Fig. \ref{fig:length-dist-org} displays the original distribution, while Fig. \ref{fig:length-dist-pred} shows the predicted distribution. Notably, the shapes of the distributions are remarkably similar and closely aligned. However, it is evident that the predicted distribution exhibits reduced variability in scanpath lengths compared to the original distribution. It is important to emphasize that our model was not trained on this dataset. That makes the close resemblance of the distributions a more meaningful achievement. This suggests that our model can generalise well on unseen data and can effectively predict scanpath lengths in different scenarios.

To summarize, this ablation study highlights the importance of the fixations selector component in our model. As the model's performance is affected by the absence of the fixations selector. it is critical for achieving the best overall performance in scanpath prediction.

\subsection{Quantitative evaluation of the generative network}
\label{subsec:gener-results}

\begin{table}[h!]
\begin{center}
\scalebox{0.9}{
\begin{tabular}{ c c c c c c }
\hline

\textbf{Metric} &   \textbf{0.0} &\textbf{0.8} &\textbf{2.0}  & \textbf{5.3} &\textbf{10.0} \\
\hline

\textbf{NSS}  & \textbf{1.426576}  &1.175942 & 1.003903 & 0.839459 & 0.789592  \\
\hline

\textbf{Congruency}  &  \textbf{0.304958} &  0.256472 & 0.224562  & 0.193916 & 0.185359     \\
\hline

\textbf{Shape}   &  \textbf{0.972415} & 0.964157 & 0.959167 & 0.952888  & 0.951019       \\
\hline

\textbf{Direction}     &    0.537310 & 0.537695 & 0.537575 &0.539900 &\textbf{0.544388}    \\
\hline

\textbf{Length}  &  \textbf{0.945042}  &0.928527 &0.902239  &0.918545 & 0.905984 \\
\hline

\textbf{Position}     &  \textbf{0.467996} & 0.462994 &0.459152  & 0.455303 &0.453116   \\
\hline

\textbf{Mean}     &  \textbf{0.7306} & 0.7233 & 0.7145  & 0.7166 & 0.7136   \\
\hline
\end{tabular}
}

\caption{ \label{tab:ablation-temp-uni}Results of scanpath prediction on MIT1003 dataset according to temperature with uniform noise}

\end{center}

\end{table}

\begin{table}[h!]
\begin{center}
\scalebox{0.9}{
\begin{tabular}{ c c c c c c }
\hline

\textbf{Metric} & \textbf{0.0} &      \textbf{0.8} &    \textbf{2.0} &      \textbf{5.3} &   \textbf{ 10.0} \\
\hline 
\textbf{NSS} &  \textbf{0.3461} &  0.2910 &  0.2643 &  0.2562 &  0.2512\\
\hline
\textbf{Congruency} & \textbf{ 0.1077} &  0.0982 &  0.0949 &  0.0914 &  0.0921 \\
\hline
\textbf{Shape}   &  \textbf{0.9761} &  0.9685 &  0.9627 &  0.9542 &  0.9516 \\
\hline

\textbf{Direction}     &  0.5423 &  0.5423 &  0.5419 &  0.5440 &  \textbf{0.5485} \\
\hline

\textbf{Length}     &  \textbf{0.9525} &  0.9372 &  0.9256 &  0.9086 &  0.9034 \\
\hline
\textbf{Position}     &  \textbf{0.5593} &  0.5509 &  0.5467 &  0.5427 &  0.5405 \\
\hline

\textbf{Mean}    &  \textbf{0.7576} &  0.7497 &  0.7442 &  0.7374 &  0.7360 \\
\hline
\end{tabular}
}

\caption{ \label{tab:ablation-temp_gauss}Results of scanpath prediction on MIT1003 dataset according to temperature with Gaussian noise}

\end{center}

\end{table}

Table \ref{tab:ablation-temp-uni} displays the results of our scanpath prediction model on the MIT1003 dataset, with varying temperature values for uniform noise during the inference process. The table presents the evaluation scores for different temperature settings, including $0.0$, $0.8$, $2.0$, $5.3$, and $10.0$.

For the $NSS$ metric, the highest score is achieved with a temperature of $0.0$, indicating that a marignial amount of noise $\epsilon$ was added. As the temperature value increases, the $NSS$ score decreases gradually, suggesting that higher levels of noise negatively impact the model's ability to align with salient regions.

In terms of the $Congruency$ metric, a similar trend is observed. The model performs best with a temperature of $0.0$, and as the temperature increases, the $Congruency$ score decreases, signifying a reduction in spatial overlap and similarity between predicted and observed fixations.

Analyzing the individual components of the $MultiMatch$ metric (Shape, Direction, Length, and Position), it can be observed that the best performance is achieved with a temperature of $0.0$ for most of these characteristics. However, for the Direction and Position components, higher temperature values show slightly improved scores.

The overall mean score, representing the $MultiMatch$ metric's average, shows a gradual decrease with increasing temperature. It indicates that the model's performance declines as more noise is introduced during inference.

In summary, the results in Table \ref{tab:ablation-temp-uni} highlight the impact of temperature and uniform noise on the model's performance. Lower temperature values (less noise) generally lead to better alignment with salient regions and improved prediction accuracy, while higher temperature values (more noise) result in reduced performance. The findings underscore the importance of carefully choosing the temperature parameter during inference to achieve optimal results in scanpath prediction.

The results in Table \ref{tab:ablation-temp_gauss} show a similarity to the findings observed in the Salicon dataset evaluation (Table \ref{tab:ablation-temp-uni}). In both cases, introducing Gaussian noise through temperature parameterization has an impact on the model's performance, as indicated by the $NSS$ and $Congruency$ scores.

For both datasets, the best performance in terms of $NSS$ and $Congruency$ is achieved with a temperature of $0.0$, Indicating that marginal noise addition leads to the highest alignment with salient regions and observed fixations. As the temperature increases (e.g., $0.8$, $2.0$, $5.3$, and $10.0$), the model's performance gradually decreases, resulting in reduced alignment with saliency maps and observed fixations.

Additionally, the individual components of the $MultiMatch$ metric (Shape, Direction, Length, and Position) follow similar trends in both datasets. Lower temperature values generally lead to better performance in most characteristics, with some slight variations observed in the Direction component.

Moreover, the overall mean score, representing the average performance of the $MultiMatch$ metric, shows a consistent decrease with increasing temperature, indicating that Gaussian noise negatively impacts the model's prediction accuracy.

The similarity in trends between the two datasets reinforces the importance of selecting an appropriate temperature parameter during inference to achieve optimal performance and better generalize the model's predictions on the task. 

\subsection{Quantitative evaluation  painting}
\label{subsec:results-paintings}


\begin{table}[htp!]
\begin{center}
\scalebox{1}{
\begin{tabular}{ c c c}
\hline
\textbf{Model} & \textbf{NSS} & \textbf{Congruency}  \\ 
\hline
Our model (without DA)  &  1.3620  &   0.4024   \\
\hline
Our model (with DA)  &  \textbf{1.5093}  &    \textbf{0.4244}  \\
\hline
\end{tabular}}
\caption{\label{tab:lemeur_results}Results of scanpath prediction on Le Meur paintings before and after domain adaptation.}
\end{center}
\end{table}

\begin{table*}[h!]
\begin{center}
\begin{tabular}{ c c c c c c c c }
\hline
\textbf{Dataset} & \textbf{Shape } & \textbf{Direction} & \textbf{Length} & \textbf{Position} & \textbf{MM Score} & \textbf{NSS} & \textbf{Congruency}\\ 
\hline
 W/o DA &  0.9548 &  \textbf{0.6198}  & 0.9363 & 0.7872  &  0.8245 &  0.64761 &   0.1212   \\
\hline
With DA & \textbf{0.9697} & 0.6173  & \textbf{0.9384} & \textbf{0.7992}  & \textbf{0.8312}  &  \textbf{0.6534} &  0.1854    \\
\hline
\end{tabular}
\end{center}

\caption{\label{tab:results-myDS-DACompare}Results of scanpath prediction of our model on AVAtt painting datasets before and after DA.} 
\end{table*}

Table \ref{tab:lemeur_results} presents the evaluation results of our model on the Le Meur painting dataset after incorporating domain adaptation on the deterministic model. As the dataset lacks scanpaths, we used the hybrid metrics $NSS$ and $Congruency$ for the evaluation.

The $NSS$ metrics' scores exhibited a considerable increase after introducing domain adaptation, indicating that the model's predicted fixations align better with salient regions in the painting images. This improvement reflects the effectiveness of domain adaptation in enhancing the model's ability to capture visual attention patterns in the painting domain.

Moreover, we also observed an improvement in the $Congruency$ results. The $Congruency$ metric measures the spatial overlap between predicted scanpaths and observed fixations, and the improvement indicates a closer alignment between the two different datasets. This further demonstrates the effectiveness of domain adaptation in bridging the gap between natural scene images and painting images, allowing the model to generalize its performance to the painting domain more effectively.

In summary, the results in Table \ref{tab:lemeur_results} highlight the positive impact of domain adaptation on our model's performance when evaluated on the Le Meur painting dataset. The increased $NSS$ scores indicate better alignment with salient regions, while the improved $Congruency$ results demonstrate a closer match between predicted scanpaths and observed fixations. These findings underscore the significance of domain adaptation in enhancing the model's generalization ability to new and diverse image domains, such as paintings.

Table \ref{tab:results-myDS-DACompare} 
Shows the results obtained from testing the model on the AVAtt dataset \cite{Tliba_2022_CVPR} before and after applying the domain adaptation method to the determinative model.

Table \ref{tab:results-myDS-DACompare} presents a comparison of our model's scanpath prediction results on the AVAtt dataset before and after domain adaptation (DA). Initially, the model showed moderate performance in various metrics, including Shape, Direction, Length, Position, MM Score, NSS, and Congruency. However, after applying domain adaptation, there was a noticeable improvement in the model's performance. The Shape, Length, Position, MM Score, and NSS metrics significantly increased, while Congruency showed substantial improvement. These results demonstrate the effectiveness of domain adaptation in enhancing the model's accuracy in predicting scanpaths and saliency values, making it more reliable and suitable for real-world applications.



\begin{table*}[h!]
\begin{center}
\begin{tabular}{ c c c c c c c c }
\hline
\textbf{Dataset} & \textbf{Shape } & \textbf{Direction} & \textbf{Length} & \textbf{Position} & \textbf{MM Score} & \textbf{NSS} & \textbf{Congruency}\\ 
\hline
 AVAtt Dataset & 0.9883 &  0.5046 & 0.9767 & 0.3718  &  0.7104 &  0.6323 &   0.1131   \\
\hline
 Le Meur Dataset & - & -  & - & -  & -  &  1.2145 &  0.3523    \\
\hline
\end{tabular}
\end{center}

\caption{\label{tab:results-my DS}Results of scanpath prediction of our stochastic model with Gaussian noise on painting datasets.}
\end{table*}

\subsection{Qualitative results}
\label{subsec:qualitative}

\begin{figure*}
    \centering
    \includegraphics[width=1\textwidth]{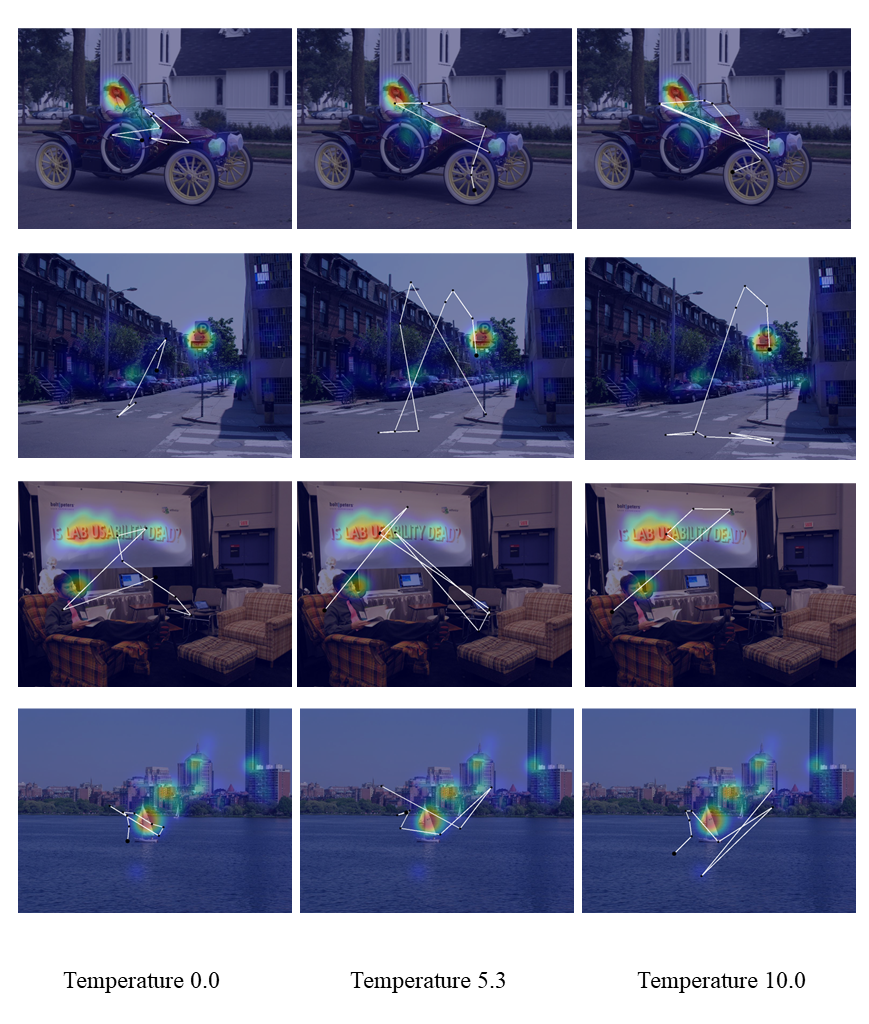}
    \caption{Qualitative predictions on MIT1003.}
    \label{fig:qual_mit}
\end{figure*}

\begin{figure*}
    \centering
    \includegraphics[width=1\textwidth]{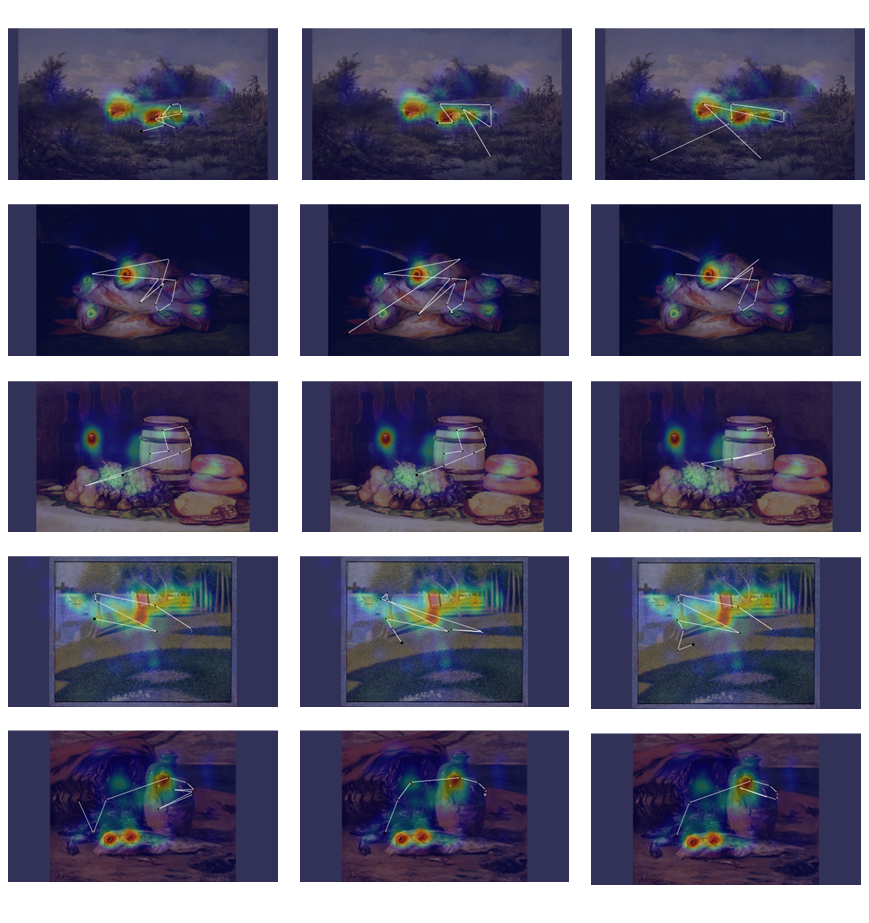}
    \caption{Qualitative predictions on Le Meur dataset.}
    \label{fig:qual-lemeur}
\end{figure*}

\begin{figure*}
    \centering
    \includegraphics[width=1\textwidth]{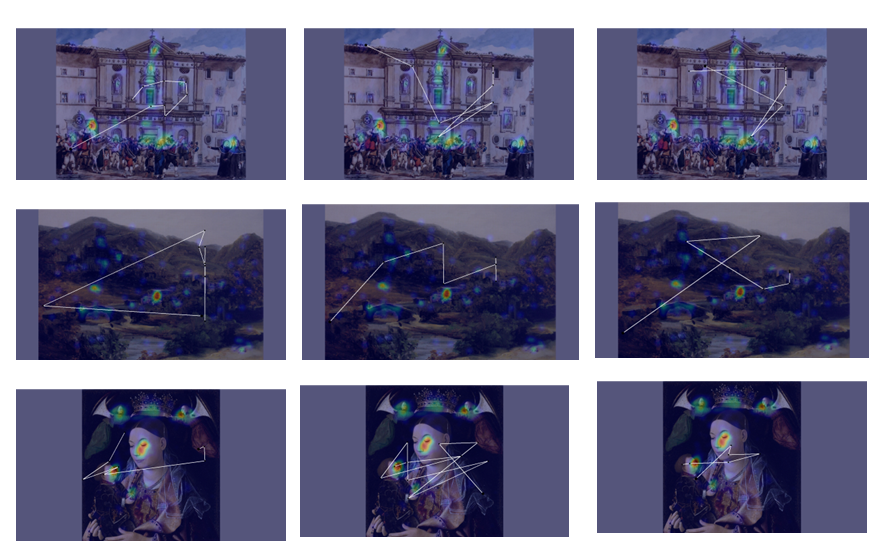}
    \caption{Qualitative predictions on AVAtt dataset.}
    \label{fig:qual-ava}
\end{figure*}

In Figure \ref{fig:qual_mit}, we provide a qualitative visualization of the results generated by our model. The figure showcases the predicted scanpaths overlaid on the image along with the ground-truth saliency map. Each column corresponds to a sample of results obtained using different temperatures as discussed in previous sections.

The depicted results demonstrate that our model is capable of efficiently and effectively predicting plausible scanpaths from input images. Notably, as we increase the temperature parameter, the influence of the central bias decreases. Consequently, scanpaths are predicted using higher temperatures spread over a larger area of the stimulus. On the other hand, scanpaths predicted with lower temperatures tend to have more concentrated fixations near the center. This observation highlights the temperature's impact on the spread and density of predicted scanpaths, offering valuable insights into the behavior of our model under varying conditions. Overall, these qualitative results affirm the model's ability to generate meaningful and accurate scanpaths from visual stimuli, making it a promising approach for scanpath prediction tasks.

In Figure \ref{figure:QualtDA}, we present a visual and qualitative illustration of the domain adaptation process for scanpath generation on a new domain of paintings. The top row shows the scanpaths generated before domain adaptation, while the bottom row displays the scanpaths obtained after applying domain adaptation.

Before domain adaptation, the scanpaths exhibit short saccadic lengths and concentrate fixations on a limited region, which might not necessarily correspond to the salient areas. This behavior can be attributed to the influence of painting styles, which can significantly impact the scanpath generation.

However, after domain adaptation, we observe a significant improvement in the scanpaths' quality. The adapted scanpaths cover wider regions and correlate strongly with the underlying saliency maps depicted in the last row. This improvement validates the effectiveness of the domain adaptation method in enhancing the model's ability to generate scanpaths that align better with the saliency patterns in the new domain of paintings.

Overall, the visual comparison showcases the benefits of domain adaptation, enabling the model to produce more accurate and contextually relevant scanpaths in the presence of new and diverse painting styles.

In Figure \ref{fig:qual-lemeur} and Figure \ref{fig:qual-ava}, we showcase samples of results obtained from the Le Meur and AVAtt datasets, respectively. These visualizations are similar to the previous results for MIT1003 and demonstrate the performance of our model across various art movements and styles.

From the images, it is evident that our model can generate multiple reasonable scanpaths for each painting. The scanpaths effectively capture the salient regions of the artworks, and the predicted fixations align well with the important elements within the paintings.

Moreover, our model showcases its versatility by performing well across various art movements and styles, ranging from the Renaissance and Baroque movements to ancient Chinese ink paintings. This highlights the robustness and adaptability of our model, allowing it to handle diverse artistic representations and generate accurate scanpaths across different visual domains.

In summary, the qualitative results from both the Le Meur and AVAtt datasets further validate the effectiveness and generalizability of our model in predicting plausible and saliency-driven scanpaths for various paintings and art styles.

\section{Conclusion}
\label{sec:conclusion}

In this paper, we presented a novel deep learning architecture for predicting scanpaths in 2D images, with a specific focus on adapting the approach to painting images, which represent a fundamental segment of cultural heritage. Our model comprises a feature extractor, learnable task-oriented prior maps, and a fixation selection module to introduce stochasticity in scanpath length prediction. We also introduce a temperature-modulated mechanism to introduce variability to predicted scanpaths in accordance with the subjective nature of scanpaths. 

We evaluated the proposed model against state-of-the-art approaches and found that it outperforms them, producing the best results in scanpath prediction. Additionally, we demonstrated the effectiveness of unsupervised domain adaptation in adapting the model from the natural scenes domain to the paintings domain, resulting in highly relevant scanpath predictions for paintings.

However, our model has limitations, as it currently lacks the ability to predict the duration of fixations. In future work, we plan to address this limitation by incorporating a time prediction module. 

Overall, our model's domain adaptation capability opens doors to further applications in the context of cultural heritage understanding, such as analyzing monuments, architectures, statues, and artificial 3D images for virtual museum visits and other heritage-related tasks.

\section*{Acknowledgment}
Funded by the TIC-ART project, Regional fund (Region Centre-Val de Loire)

{\small
\bibliographystyle{ieee_fullname}
\bibliography{egbib}
}

\end{document}